\documentclass{article} % For LaTeX2e
\usepackage{iclr2027_conference,times}

\usepackage{amsmath,amsfonts,bm}

\def\eqref#1{equation~\ref{#1}}
\def\1{\bm{1}}

\DeclareMathAlphabet{\mathsfit}{\encodingdefault}{\sfdefault}{m}{sl}
\SetMathAlphabet{\mathsfit}{bold}{\encodingdefault}{\sfdefault}{bx}{n}

\usepackage{hyperref}
\usepackage{url}
\usepackage{enumitem}
\usepackage{graphicx}
\usepackage{booktabs}
\usepackage{tabularx}
\usepackage{array}
\usepackage{wrapfig}
\usepackage{multirow}

\newcolumntype{Y}{>{\raggedright\arraybackslash}X}

\title{CoRe-VLA: Preserving Cross-View Coordination in VLAs under Camera Shifts}

\author{%
\textbf{Tianhang Pan\textsuperscript{1}, Xuanhao Wang\textsuperscript{1}, Yiwen Pang\textsuperscript{1}, Bo Zhou\textsuperscript{1},}\\
\textbf{Jun Yang\textsuperscript{1,2}, Min-Ling Zhang\textsuperscript{1} \& Shimin Di\textsuperscript{1}}\\[2pt]
\textsuperscript{1}Southeast University, Nanjing, China\\
\textsuperscript{2}National Center of Technology Innovation for EDA, Nanjing, China\\
\texttt{\{tianhang.pan,shimin.di\}@seu.edu.cn}
}

\iclrfinalcopy 
\begin{document}

\maketitle
\lhead{}
\renewcommand{\headrulewidth}{0pt}

\begin{abstract}
VLAs combine pretrained vision-language representations with action generation to enable language-guided control across diverse tasks, becoming a mainstream paradigm in embodied intelligence.
However, multiple studies have reported VLA's substantial declines in task success under camera shifts, revealing a key vulnerability that limits reliable deployment.
To address this vulnerability, existing methods collect paired observations of the same scene from different viewpoints to fine-tune the VLA or train visual adaptation modules.
Unfortunately, they require additional data collection and VLA training costs.
In this paper, we first identify \emph{cross-view coordination breakdown} under external camera shifts: the robot may rely too heavily on wrist-view cues and consequently execute subtasks in the wrong order when losing global view.
Motivated by this, we propose CoRe-VLA, a plug-and-play framework requiring neither additional multi-view data collection nor VLA fine-tuning, which can incorporate with exsiting VLAs.
It reconstructs a scene point cloud and renders the observation from the VLA's training viewpoint to restore cross-view coordination.
In CoRe-VLA, Render-to-Camera (R2C) Restoration reduces rendering-induced visual degradation, while Execution-Trajectory-Conditioned Alignment (ETCA) reduces robot idle time and mitigates motion conflicts during asynchronous execution.
Experiments on 5 real-robot tasks, LIBERO-100 and LIBERO-Plus demonstrate CoRe-VLA substantially improves task success across mainstream VLAs under camera shifts.
For example, CoRe-VLA raises $\pi_{0.5}$'s success rate from 13.3\% to 83.3\% at a 1.6m camera shift in real-robot environment. 
\end{abstract}

\section{Introduction}
\label{sec:1}
\begin{figure}
    \centering
    \includegraphics[width=1\linewidth]{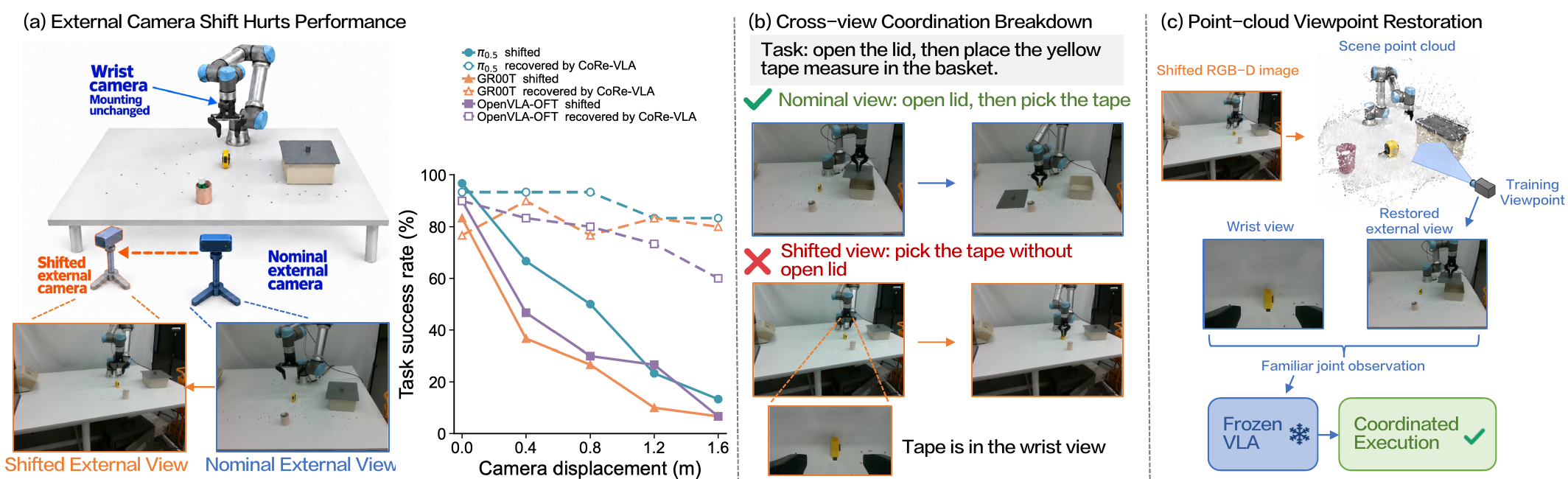}
    \caption{
    (a) The external camera shifts while the wrist-camera mounting remains unchanged. The accompanying curves compare task success rates on 5 real-robot tasks with and without CoRe-VLA.
    (b) Under the shifted view, the robot grasps the tape measure visible in the wrist view while the basket lid remains closed, violating the required subtask order.
    (c) The core idea of CoRe-VLA: using point-cloud reconstruction and re-rendering to restore the external observation and recover cross-view coordination.
}
    \label{fig:fig1}
    \vspace{-15pt}
\end{figure}

Embodied intelligence extends AI from understanding digital information to interacting with the physical world, enabling robots to perform useful tasks across a wide range of scenarios.
Vision-language-action (VLA) models have emerged as a prominent approach to this goal, translating visual observations and language instructions into robot actions with strong performance across diverse manipulation tasks.
OpenVLA~\citep{kim2024openvla} is a pioneering work that combines pretrained vision-language representations with large-scale robot demonstrations to enable generalist robotic manipulation.
$\pi_0$~\citep{black2024pi0} combines pretrained vision-language representations with flow matching for dexterous robot control, while $\pi_{0.5}$~\citep{intelligence2025pi} improves generalization to unfamiliar environments through co-training on heterogeneous data.
GR00T~\citep{bjorck2025gr00t} combines real-robot trajectories, human videos, and synthetic data to support generalist manipulation across robot embodiments.
These mainstream open-source VLAs provide an accessible foundation for developing general-purpose robotic systems.
Despite these advances, reliable deployment of VLAs remains challenging, particularly because VLAs are highly sensitive to camera shifts.
Such shifts are common and often difficult to avoid in practice, as external-camera placement varies across deployment environments and workspace configurations.
Prior studies~\citep{fei2026libero, heo2026anycamvla} have reported substantial performance degradation under camera shifts in simulation and real-robot settings.
As shown in Fig.~\ref{fig:fig1}(a), our evaluation on 5 real-robot tasks further demonstrates this vulnerability: mainstream VLAs' task success rates decline substantially as external-camera shift increases.
For example, a shift of 0.4~m reduces $\pi_{0.5}$'s success rate from 96.7\% to 66.7\%.
%
%These findings highlight the gap between strong manipulation capabilities under familiar viewpoints and reliable execution after camera repositioning.
These findings highlight the gap in maneuverability between familiar viewpoints and after camera repositioning.

Existing approaches try to address viewpoint sensitivity through VLA fine-tuning or training a visual adaptation module, introducing additional data preparation and VLA training costs.
Cross-View Action Consistency~\citep{huang2026cross} fine-tunes VLAs for consistent action-flow predictions across viewpoints.
AnyCamVLA~\citep{heo2026anycamvla} and GS-VLA~\citep{park2026gs} instead restore familiar training views through pretrained novel-view synthesis and learned Gaussian-based canonicalization, respectively, while keeping the VLA frozen.
Their visual adaptation modules depend on multi-view supervision for training or domain adaptation, which must be collected in real robotic environment, further increasing the already substantial cost of robot data collection.
These limitations raise a natural question: how to imporve existing VLAs' viewpoint robustness without additional multi-view data collection or VLA retraining?
To guide the design of such an approach, we systematically analyze how these shifts affect robot execution.
Our real-robot experiments reveal \emph{cross-view coordination breakdown}: when the external camera shifts, the robot may rely too heavily on wrist-view cues and consequently execute subtasks in the wrong order.
For example, in Fig.~\ref{fig:fig1}(b), under the shifted view, the robot grasps the tape measure visible in the wrist view while the basket lid remains closed, violating the required subtask order.
Although the grasp succeeds, the action does not account for the lid-closed state visible in the external view.
This finding motivates us to explore how to restore the nominal external view for the VLA to prevent cross-view coordination breakdown.
Point-cloud reconstruction and re-rendering provide a simple and effective way to achieve this goal, as illustrated in Fig.~\ref{fig:fig1}(c).
A scene point cloud is reconstructed from RGB-D observations and rendered at the training-time external camera pose.
This geometric approach requires neither additional multi-view demonstration collection nor VLA training or fine-tuning, making it suitable for integration with existing models. 
However, translating this idea into reliable robot control presents two challenges: 
(i) rendering degrades texture, color, and sharpness, and efficiently correcting these combined artifacts remains challenging, (ii) reconstruction and rendering increase latency in VLA inference, slowing task execution.
To address above challenges, we propose Coordination Recovery for Vision-Language-Action Models (\textbf{CoRe-VLA}), a framework built on point-cloud reconstruction and viewpoint transformation. 
To reduce rendering artifacts, Render-to-Camera (\textbf{R2C}) Restoration uses a lightweight NAFNet~\citep{chen2022simple} trained on degraded original image pairs constructed from existing demonstration videos.
Perceptual, color, and pixel losses jointly guide texture recovery, color correction, and detail reconstruction.
To mitigate the additional latency from rendering and restoration, Execution-Trajectory-Conditioned Alignment (ETCA) is introduced for asynchronous control.
Because the robot continues moving during inference, ETCA matches each newly predicted action chunk with the recorded execution trajectory to identify where execution should continue.
This asynchronous control strategy substantially offsets the increase in task execution time introduced by CoRe-VLA while automatically reducing motion conflicts under high inference latency.
Together, these components make CoRe-VLA a plug-and-play method for improving viewpoint robustness, requiring neither additional multi-view data collection nor VLA retraining.
The main contributions are listed:
\begin{enumerate}[leftmargin=*, labelindent=0pt]
    \item We identify cross-view coordination breakdown as a key failure mechanism in VLAs under external-camera shifts.
    \item We propose ETCA, an asynchronous control strategy that can be applied to all robot policies to reduce task execution time and automatically mitigates motion conflicts.
    \item We propose CoRe-VLA, a plug-and-play method that integrates with existing VLAs to preserve cross-view coordination without additional data collection or VLA retraining. Experiments show that CoRe-VLA substantially improves the viewpoint robustness of mainstream VLAs.
\end{enumerate}

\section{Related Work}
\noindent
\textbf{VLAs and Viewpoint Robustness.}
The core idea of VLA models is to combine a pretrained VLM with an action generation module~\citep{kim2024openvla, intelligence2025pi, bjorck2025gr00t}.
Because the VLM directly supplies task-relevant visual representations for downstream control, perturbations to visual inputs can disrupt action generation, and substantially reduce task success rates.
LIBERO-Plus~\citep{fei2026libero} identifies camera-viewpoint perturbations as a major vulnerability, while VLA-Arena~\citep{zhang2025vla} reports sharp degradation when camera shifts are introduced into its visual perturbation protocol.
Li et al.~\citep{li2026vla} further attribute viewpoint-induced failures primarily to spatial representation misalignment and show that lightweight visual adaptation can recover performance.
\noindent
\textbf{World Action Models}
World action models (WAMs) leverage pretrained video models to learn robot control while predicting how the scene evolves.
DreamZero~\citep{ye2026world} jointly models video and actions to improve generalization to unseen tasks and environments, while Cosmos Policy~\citep{kim2026cosmos} predicts actions, future states, and values for control and planning.
However, evaluations on LIBERO-Plus~\citep{fei2026libero} and RoboTwin 2.0-Plus~\citep{zhang2026world} show that WAMs remain sensitive to camera shifts despite their robustness to lighting and noise perturbations~\citep{zhang2026world}.
Selective Cross-View Consistency~\citep{huang2026selective} addresses this limitation through training on same-state viewpoint pairs, improving performance on held-out extrapolation viewpoints.
These findings suggest that adopting a WAM alone does not resolve viewpoint sensitivity, which still requires targeted treatment.
\noindent
\textbf{Point Cloud Reconstruction}
Point cloud can provide explicit geometry for rendering scene observations from arbitrary viewpoints.
Relevant methods for reconstructing point cloud from image observations can be grouped into three categories: neural point-based rendering, Gaussian splatting, and direct RGB-D reprojection.
Neural point-based methods, represented by NPBG~\citep{aliev2020neural}, learn point descriptors and a rendering network, but require scene-specific fitting that complicates deployment in changing embodied manipulation scenes.
Gaussian-based methods such as 3D Gaussian Splatting~\citep{kerbl2023gaussian} enable real-time rendering, yet their original pipelines require multi-view capture and per-scene optimization rather than immediate reconstruction from the current observation.
In contrast, direct RGB-D reprojection reconstructs and renders the observed geometry without scene-specific optimization, enabling efficient real-time viewpoint restoration~\citep{schops2017realtime}.
However, its reconstruction accuracy depends strongly on depth accuracy and observation coverage.
Depth errors, missing surfaces, and imperfect color alignment make texture loss, color distortion, and reduced sharpness difficult to avoid in practice~\citep{zhou2014color,fink2023livenvs}.
Online neural rendering methods such as LiveNVS~\citep{fink2023livenvs} improve visual quality through multi-view feature fusion, but differ from restoring each current observation independently.
Jointly reducing these degradations from a single RGB-D observation while maintaining low processing latency remains challenging, motivating our lightweight R2C restoration module.

\section{Understanding VLA Failures under External-Camera Shifts}
\label{sec:3}

\begin{table}[t]
\centering
\caption{Real-world task suite used to diagnose VLA failures under external-camera shifts. T1 evaluates basic manipulation, T2--T3 long-horizon execution, and T4--T5 spatial reasoning.}
\label{tab:real_robot_tasks}
\tiny
\setlength{\tabcolsep}{4pt}
\renewcommand{\arraystretch}{1.12}
\begin{tabular*}{\linewidth}{@{\extracolsep{\fill}}cll@{}}
\toprule
\textbf{ID} & \textbf{Category} & \textbf{Task instruction} \\
\midrule
T1 & Basic & Pick up the green glue stick and place it in the pink cup. \\
T2 & Long-horizon & Place the yellow tape measure and the pink cup into the basket in sequence. \\
T3 & Long-horizon & Open the lid, then pick up the yellow tape measure and place it in the basket. \\
T4 & Spatial & Pick up the green glue stick and place it on the side of the pink cup farther from the basket. \\
T5 & Spatial & Pick up the object closest to the basket and place it inside. \\
\bottomrule
\end{tabular*}
\end{table}

\begin{figure}[t]
    \vspace{-10pt}
    \centering
    \begin{minipage}[b]{0.25\linewidth}
        \centering
        \includegraphics[width=\linewidth]{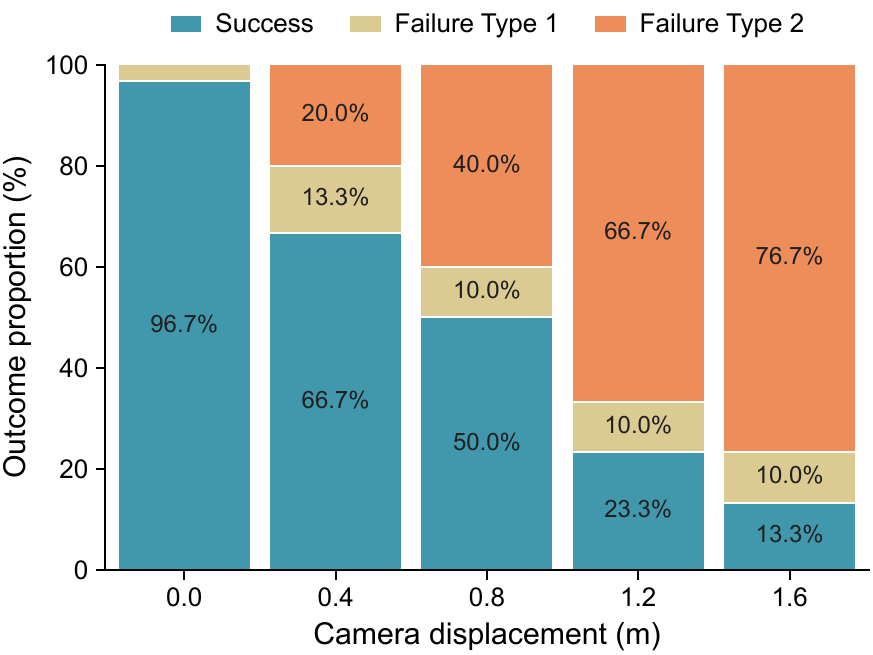}
        \par\vspace{-2pt}
        {\small (a) $\pi_{0.5}$}
    \end{minipage}\hfill
    \begin{minipage}[b]{0.25\linewidth}
        \centering
        \includegraphics[width=\linewidth]{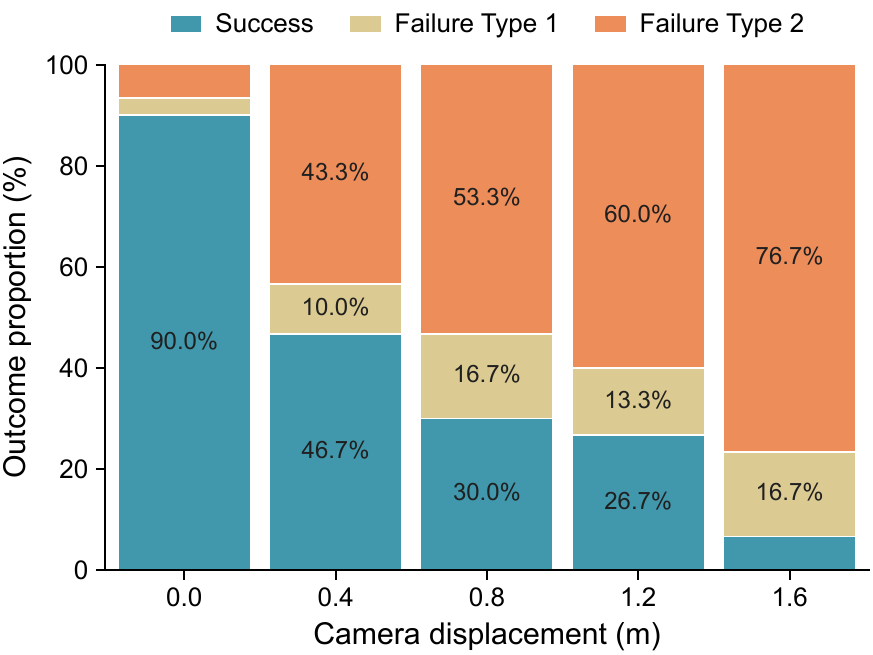}
        \par\vspace{-2pt}
        {\small (b) OpenVLA-OFT}
    \end{minipage}\hfill
    \begin{minipage}[b]{0.25\linewidth}
        \centering
        \includegraphics[width=\linewidth]{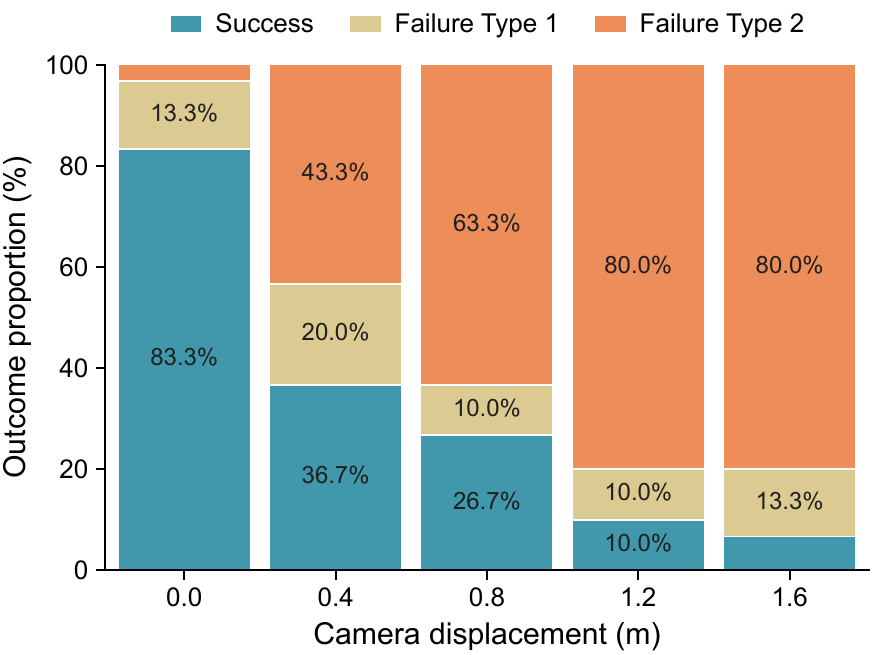}
        \par\vspace{-2pt}
        {\small (c) GR00T}
    \end{minipage}
    \vspace{-10pt}
    \caption{
    Outcome distributions on five real-robot tasks under external-camera shifts for (a) $\pi_{0.5}$, (b) OpenVLA-OFT, and (c) GR00T.
    Failure Type 1 denotes failure to complete any meaningful subtask, while Failure Type 2 denotes local manipulation inconsistent with the global task state.
    }
    \label{fig:fig2}
    \vspace{-15pt}
\end{figure}

\textbf{Real-World Diagnostic Setup.}
To investigate how to design a viewpoint-robust method that require neither multi-view data collection nor VLA training, we systematically analyze how external-camera shifts affect robot execution in multi-camera VLA systems. 
As a prerequisite, we construct a real-world benchmark using a UR7e robot equipped with an external RGB-D camera and a wrist-mounted camera.
Detailed task definitions are provided in Table~\ref{tab:real_robot_tasks}.
The tasks cover basic manipulation (T1), long-horizon execution (T2-T3), and spatial reasoning (T4-T5).
Full task instructions and success criteria are provided in Appendix~\ref{tab:app_success_criteria}.
Three mainstream VLA models are evaluated: $\pi_{0.5}$~\citep{intelligence2025pi}, OpenVLA-OFT~\citep{kim2025fine}, and GR00T~\citep{bjorck2025gr00t}.
Each model is fine-tuned on 100 demonstrations per task collected at the nominal external-camera pose.
During evaluation, the wrist-camera mounting remains fixed relative to the end effector, while the external camera is displaced by 0.4, 0.8, 1.2, or 1.6 m.
At each distance, three camera positions are randomly sampled, with two fixed initial object layouts per task.
Each task execution is classified as successful or as one of two failure types:
Type 1 indicates failure to complete any meaningful subtask, such as moving irregularly or failing to grasp an object;
Type 2 indicates a successful local manipulation that is inconsistent with the global task state.
For example, the robot may grasp the tape measure before opening the lid in task 3.
This taxonomy separates local manipulation failures from failures in selecting task-appropriate actions.
The same platform and tasks are reused to evaluate CoRe-VLA in Sec.~\ref{sec:5}.
\textbf{Quantitative and Qualitative Failure Analysis.}
Fig.~\ref{fig:fig2} (a) shows the outcome distributions across external-camera shifts. 
Across the evaluated models, task success declines as displacement increases, while task-state inconsistency (Failure Type 2) becomes the dominant failure mode. 
For example, the success rate of $\pi_{0.5}$ drops from 98.0\% at the nominal viewpoint to 13.3\% at a displacement of 1.6 m, while Type 2 increases from 0\% to 76.7\% of all trials. 
In contrast, local execution failures (Failure Type 1) account for only 10.0\% to 13.3\% of trials under nonzero displacements. 
These results suggest that viewpoint shifts substantially impair the selection of task-appropriate actions, even when local manipulation remains feasible.
In addition to the example discussed in Sec.~\ref{sec:1}, Task 5 in Fig.~\ref{fig:app_execution_examples} in Appendix is also a typical example.
The task requires selecting the object closest to the basket based on the global spatial relationships captured by the external camera.
After the camera shifts, the robot instead grasps the cup visible in the wrist view, failing to coordinate this local cue with the global spatial constraint.
By contrast, the task succeeds when the camera remains at its nominal position.
Together, these observations provide behavioral evidence of \emph{cross-view coordination breakdown}: when the external camera shifts, the robot may rely too heavily on wrist-view cues and consequently execute subtasks in the wrong order.
This finding motivates us to focus our method design on restoring the nominal external view for the VLA to prevent cross-view coordination breakdown.

\section{Methodology}

\begin{figure}
    \centering
    \includegraphics[width=1\linewidth]{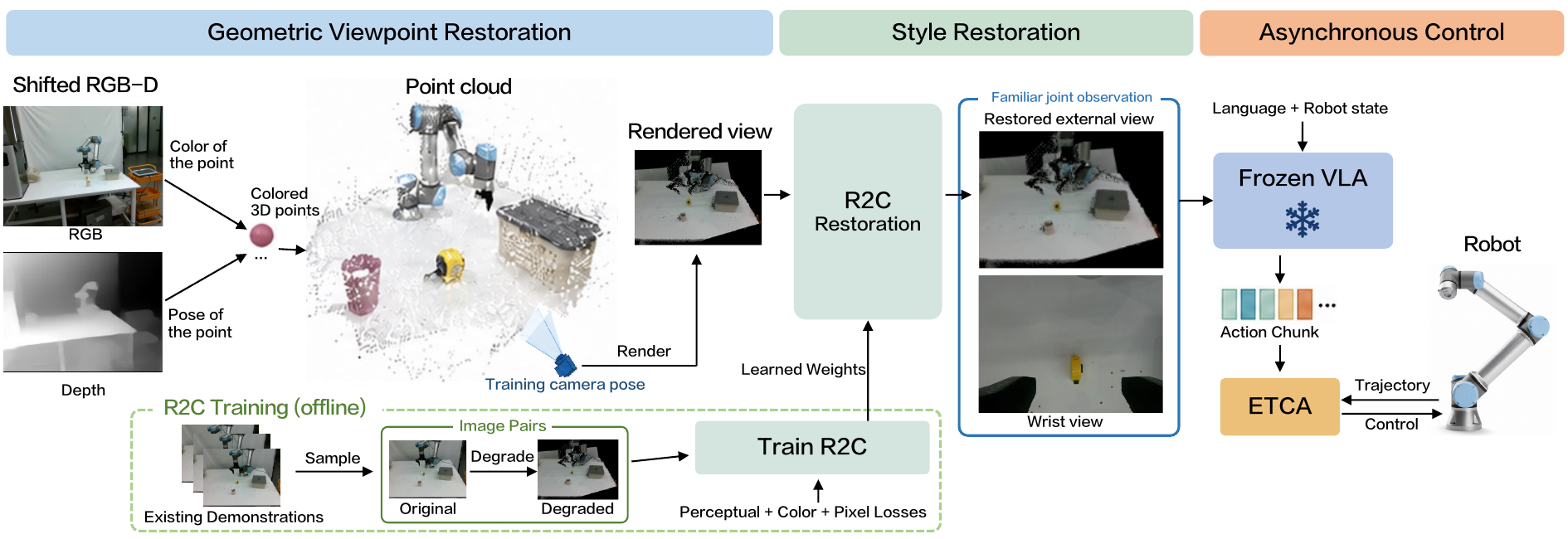}
    \caption{
    Overview of CoRe-VLA, comprising three components: geometric viewpoint restoration, R2C for image restoration, and ETCA for asynchronous control.
    }
    \label{fig:fig3}
    \vspace{-10pt}
\end{figure}

In this section, we present Coordination Recovery for Vision-Language-Action Models (CoRe-VLA).
As shown in Fig.~\ref{fig:fig3}, CoRe-VLA comprises three components: Geometric Viewpoint Restoration, Render-to-Camera (R2C) Restoration, and Execution-Trajectory-Conditioned Alignment (ETCA).

\subsection{Geometric Viewpoint Restoration}
As outlined in before, we adopt point-cloud reconstruction and re-rendering based on direct RGB-D reprojection to restore the nominal external view for the VLA.
This strategy offers two advantages: (i) it provides a simple and efficient way to restore the training-view geometry, (ii) it requires neither training a view-transformation model nor collecting paired multi-view data.
Specifically, given an aligned RGB-D observation $(G_s,D_s)$ from the shifted external camera, our goal is to synthesize an image $G_p$ at the training camera pose.
This process uses the source and target camera intrinsics, $K_s$ and $K_0$, and the known rigid transformation $(R_{0\leftarrow s},t_{0\leftarrow s})$ between their coordinate systems.
For each pixel $\mathbf{u}$ in shifted external view with valid depth, its homogeneous coordinates $\tilde{\mathbf{u}}$ are first back-projected into 3D using $D_s$ and $K_s$.
The resulting point is transformed into the training camera coordinate system and projected onto the target image plane:
\begin{equation}
\begin{aligned}
\mathbf{x}_0(\mathbf{u})
&=
R_{0\leftarrow s}
\left[D_s(\mathbf{u})K_s^{-1}\tilde{\mathbf{u}}\right]
+t_{0\leftarrow s},\\
\mathbf{v}
&=
\pi\!\left(K_0\mathbf{x}_0(\mathbf{u})\right),
\end{aligned}
\end{equation}
where $\mathbf{x}_0(\mathbf{u})$ denotes the transformed 3D point and $\pi([x,y,z]^\top)=[x/z,y/z]^\top$ performs perspective division.
Each point retains its source color $G_s(\mathbf{u})$, and the transformed colored point cloud is rasterized to produce $G_p$.
By presenting the current scene in the training-view geometry, $G_p$ provides a basis for restoring coordination between global task context and wrist-level visual cues.

\subsection{Render-to-Camera Restoration}
\label{sec:r2c}
Geometric viewpoint restoration recovers the training-view geometry, but rendering from a sparse and imperfectly reconstructed point cloud can introduce loss of material and texture detail, color distortion, and reduced sharpness into $G_p$.
Jointly reducing these three degradations from a single RGB-D observation while maintaining low processing latency is challenging. 
Non-learning method pipelines require manual tuning and can accumulate errors, while off-the-shelf deep-learning methods can not address these artifacts jointly.
Multi-step diffusion methods can also incur substantial latency.
These considerations motivate a lightweight restoration module with supervision specifically tailored to rendering-induced degradations.
In our proposed Render-to-Camera (R2C) Restoration, a lightweight NAFNet-based network~\citep{chen2022simple} is trained with perceptual, color, and pixel losses to jointly promote texture recovery, color consistency, and fine-detail reconstruction.
At inference time, the network $f_{\theta}$ maps the rendered image $G_p$ to the restored external image $G_r$.
R2C is trained on image pairs constructed from existing 2D task demonstration videos, avoiding additional multi-view data collection (Fig.~\ref{fig:fig3}).
Each original frame $G_0$ is paired with a degraded counterpart $\tilde{G}_0$ that reproduces rendering artifacts while preserving the scene state and viewpoint.
Construction details are provided in Appendix~\ref{app:r2c_pairs}.
For a restored prediction $G_r=f_{\theta}(\tilde{G}_0)$ and its original target $G_0$, the joint training objective is
\begin{equation}
\mathcal{L}_{\mathrm{R2C}}
=
\lambda_{\mathrm{per}}\mathcal{L}_{\mathrm{per}}
+
\lambda_{\mathrm{col}}\mathcal{L}_{\mathrm{col}}
+
\lambda_{\mathrm{pix}}\mathcal{L}_{\mathrm{pix}},
\end{equation}
where
\begin{equation}
\begin{aligned}
\mathcal{L}_{\mathrm{per}}
&=
\sum_{\ell}\frac{a_{\ell}}{N_{\ell}}
\left\|\phi_{\ell}(G_r)-\phi_{\ell}(G_0)\right\|_1,\\
\mathcal{L}_{\mathrm{col}}
&=
\frac{1}{C}\left(
\left\|\mu(G_r)-\mu(G_0)\right\|_1
+
\left\|\sigma(G_r)-\sigma(G_0)\right\|_1
\right)
+
\frac{\rho}{N}\left\|G_r-G_0\right\|_1,\\
\mathcal{L}_{\mathrm{pix}}
&=
\frac{1}{N}\sum_{j=1}^{N}
\sqrt{\left(G_r[j]-G_0[j]\right)^2+\epsilon^2}.
\end{aligned}
\end{equation}
Here, $\phi_{\ell}$ denotes frozen VGG16 features extracted after ImageNet normalization, while $\mu$ and $\sigma$ are channel-wise spatial means and standard deviations.
$N=HWC$ and $N_{\ell}$ count scalar elements in the image and feature map, respectively. $C$ is the number of color channels, and $G[j]$ denotes the $j$-th scalar image element.
The Charbonnier penalty uses $\epsilon=10^{-3}$.
Only the restoration network is updated, while the VLA remains frozen.

\subsection{Execution-Trajectory-Conditioned Alignment}
Although geometric viewpoint restoration and R2C enable CoRe-VLA to recover a high-quality joint observation familiar to the VLA from training, the additional processing substantially increases observation-to-action latency.
Under synchronous control, the robot must remain idle until the next action chunk is available.
Asynchronous inference and execution offer a practical way to reduce this idle time, allowing the robot to continue executing its current action chunk while the next is computed.
However, since the robot continues moving during inference, the incoming chunk generally cannot be executed from its beginning. 
Instead, a method is needed to determine where execution should begin within the incoming chunk to avoid motion conflicts. 
Such method is particularly important under high and variable latency.
\begin{wrapfigure}{r}{0.35\textwidth}
    \centering
    \includegraphics[width=\linewidth]{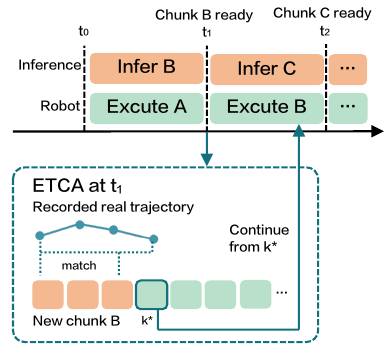}
    \caption{ETCA matches recorded robot motion to a new action chunk
    and selects where execution resumes.}
    \label{fig:etca}
\end{wrapfigure}
To address the challenge, we propose Execution-Trajectory-Conditioned Alignment (ETCA).
It treats action-chunk switching as a trajectory-matching problem: it identifies the prefix of the incoming chunk that best matches the motion already executed during inference, allowing execution to continue from the remaining suffix.
The matching considers both the overall trajectory and its terminal joint state, accounting for execution history and the robot's current configuration.
Specifically, let $\mathbf{Q}^{\mathrm{real}}=[\mathbf{q}^{r}_{0},\ldots,\mathbf{q}^{r}_{M-1}]$ denote the measured joint states recorded between observation capture and chunk arrival, and $\mathbf{A}=[\mathbf{a}_{0},\ldots,\mathbf{a}_{N-1}]$ the incoming action chunk.
For each candidate alignment index $k\in\mathcal{K}$, the predicted arm-joint prefix $[\mathbf{a}^{q}_{0},\ldots,\mathbf{a}^{q}_{k}]$ is linearly resampled to $M$ states, denoted by $\widetilde{\mathbf{A}}^{(k)}$, to enable comparison with the recorded trajectory.
The path and terminal joint-state errors are
\begin{equation}
\begin{aligned}
d_{\mathrm{path}}(k)
&=
\frac{1}{M}
\sum_{m=0}^{M-1}
\left\|
\mathbf{q}^{r}_{m}
-
\widetilde{\mathbf{a}}^{(k)}_{m}
\right\|_{2},\\
d_{\mathrm{end}}(k)
&=
\left\|
\mathbf{q}^{r}_{M-1}
-
\mathbf{a}^{q}_{k}
\right\|_{2}.
\end{aligned}
\end{equation}
ETCA selects the best-matching prefix by minimizing these errors, with motion-direction consistency and elapsed time providing auxiliary guidance:
\begin{equation}
k^{\star}
=
\arg\min_{k\in\mathcal{K}}
\left[
d_{\mathrm{path}}(k)
+
\lambda_{\mathrm{end}}d_{\mathrm{end}}(k)
+
\lambda_{\mathrm{dir}}d_{\mathrm{dir}}(k)
+
\lambda_{\mathrm{time}}|k-k_{\mathrm{time}}|
\right],
\end{equation}
where $d_{\mathrm{dir}}$ penalizes inconsistent motion directions and $k_{\mathrm{time}}$ is an alignment index estimated from elapsed inference time.
A match is accepted only if its path and terminal errors satisfy predefined thresholds and it is sufficiently distinct from competing candidates.
% 需要具体说一下计算方法（可以放到附录中）
Execution then continues from the remaining suffix with a short blending transition.
Otherwise, the chunk is rejected and inference restarts from the latest observation.

\section{Experiment}
\label{sec:5}
\subsection{Experimental Setup}
We evaluate CoRe-VLA on the 5 real-robot tasks introduced in Sec.~\ref{sec:3}, as well as LIBERO~\citep{liu2023libero} and LIBERO-Plus~\citep{fei2026libero} in simulation.
Three mainstream VLAs are included in our experiments: $\pi_{0.5}$~\citep{intelligence2025pi}, OpenVLA-OFT~\citep{kim2024openvla}, and GR00T~\citep{bjorck2025gr00t}.
The VLA parameters remain frozen when applying CoRe-VLA.
Our experiments focus on camera position shifts rather than isolated orientation changes, which can be compensated for by 2D image transformations such as rotation when task-relevant objects remain visible.
In all experiments, the camera is directed toward the tabletop center after each position shift to help keep task-relevant objects within view.
Task performance is measured by success rate and the two failure categories defined in Sec.~\ref{sec:3}: local execution failure and task-state inconsistency.
ETCA is assessed using mean completion time and mean idle time.

\begin{table}[!t]
\centering
\caption{Real-robot outcome rates (\%) across external-camera shifts. Raw and CoRe denote the shifted view and CoRe-VLA-restored view, respectively. SR denotes success rate, and F2 denotes the proportion of all trials classified as Failure Mode 2 (task-state inconsistency).}
\label{tab:main_exp_detailed}
\tiny
\setlength{\tabcolsep}{1.5pt}
\renewcommand{\arraystretch}{0.88}
\setlength{\aboverulesep}{0.15ex}
\setlength{\belowrulesep}{0.15ex}
\begin{tabular*}{\linewidth}{@{\extracolsep{\fill}}ccc*{12}{c}@{}}
\toprule
\multirow[c]{2}{*}{Shift (m)} & \multirow[c]{2}{*}{VLA} & \multirow[c]{2}{*}{Metric} & \multicolumn{2}{c}{T1} & \multicolumn{2}{c}{T2} & \multicolumn{2}{c}{T3} & \multicolumn{2}{c}{T4} & \multicolumn{2}{c}{T5} & \multicolumn{2}{c}{Overall} \\
 & & & Raw & CoRe & Raw & CoRe & Raw & CoRe & Raw & CoRe & Raw & CoRe & Raw & CoRe \\
\midrule
\multirow[c]{6}{*}{0.0} & \multirow[c]{2}{*}{$\pi_{0.5}$} & SR & 100.0 & 83.3 & 83.3 & 83.3 & 100.0 & 100.0 & 100.0 & 100.0 & 100.0 & 100.0 & 96.7 & 93.3 \\
 & & F2 & 0.0 & 0.0 & 0.0 & 0.0 & 0.0 & 0.0 & 0.0 & 0.0 & 0.0 & 0.0 & 0.0 & 0.0 \\
 & \multirow[c]{2}{*}{OpenVLA-OFT} & SR & 83.3 & 83.3 & 83.3 & 100.0 & 100.0 & 83.3 & 100.0 & 100.0 & 83.3 & 83.3 & 90.0 & 90.0 \\
 & & F2 & 0.0 & 0.0 & 16.7 & 0.0 & 0.0 & 0.0 & 0.0 & 0.0 & 16.7 & 16.7 & 6.7 & 3.3 \\
 & \multirow[c]{2}{*}{GR00T} & SR & 83.3 & 66.7 & 66.7 & 66.7 & 66.7 & 66.7 & 100.0 & 83.3 & 100.0 & 100.0 & 83.3 & 76.7 \\
 & & F2 & 0.0 & 0.0 & 16.7 & 0.0 & 0.0 & 16.7 & 0.0 & 0.0 & 0.0 & 0.0 & 3.3 & 3.3 \\
\midrule
\multirow[c]{6}{*}{0.4} & \multirow[c]{2}{*}{$\pi_{0.5}$} & SR & 33.3 & 83.3 & 83.3 & 100.0 & 33.3 & 83.3 & 83.3 & 100.0 & 100.0 & 100.0 & 66.7 & 93.3 \\
 & & F2 & 33.3 & 0.0 & 16.7 & 0.0 & 33.3 & 16.7 & 16.7 & 0.0 & 0.0 & 0.0 & 20.0 & 3.3 \\
 & \multirow[c]{2}{*}{OpenVLA-OFT} & SR & 66.7 & 66.7 & 66.7 & 100.0 & 0.0 & 66.7 & 33.3 & 83.3 & 66.7 & 100.0 & 46.7 & 83.3 \\
 & & F2 & 0.0 & 16.7 & 33.3 & 0.0 & 100.0 & 0.0 & 50.0 & 16.7 & 33.3 & 0.0 & 43.3 & 6.7 \\
 & \multirow[c]{2}{*}{GR00T} & SR & 100.0 & 66.7 & 33.3 & 100.0 & 16.7 & 100.0 & 33.3 & 83.3 & 0.0 & 100.0 & 36.7 & 90.0 \\
 & & F2 & 0.0 & 0.0 & 66.7 & 0.0 & 83.3 & 0.0 & 50.0 & 0.0 & 16.7 & 0.0 & 43.3 & 0.0 \\
\midrule
\multirow[c]{6}{*}{0.8} & \multirow[c]{2}{*}{$\pi_{0.5}$} & SR & 16.7 & 83.3 & 33.3 & 100.0 & 66.7 & 100.0 & 83.3 & 100.0 & 50.0 & 83.3 & 50.0 & 93.3 \\
 & & F2 & 33.3 & 16.7 & 66.7 & 0.0 & 33.3 & 0.0 & 16.7 & 0.0 & 50.0 & 16.7 & 40.0 & 6.7 \\
 & \multirow[c]{2}{*}{OpenVLA-OFT} & SR & 66.7 & 66.7 & 33.3 & 50.0 & 50.0 & 83.3 & 0.0 & 100.0 & 0.0 & 100.0 & 30.0 & 80.0 \\
 & & F2 & 0.0 & 0.0 & 66.7 & 33.3 & 50.0 & 16.7 & 100.0 & 0.0 & 50.0 & 0.0 & 53.3 & 10.0 \\
 & \multirow[c]{2}{*}{GR00T} & SR & 100.0 & 83.3 & 16.7 & 100.0 & 16.7 & 100.0 & 0.0 & 0.0 & 0.0 & 100.0 & 26.7 & 76.7 \\
 & & F2 & 0.0 & 0.0 & 83.3 & 0.0 & 33.3 & 0.0 & 100.0 & 83.3 & 100.0 & 0.0 & 63.3 & 16.7 \\
\midrule
\multirow[c]{6}{*}{1.2} & \multirow[c]{2}{*}{$\pi_{0.5}$} & SR & 16.7 & 50.0 & 16.7 & 100.0 & 16.7 & 83.3 & 66.7 & 100.0 & 0.0 & 83.3 & 23.3 & 83.3 \\
 & & F2 & 66.7 & 50.0 & 83.3 & 0.0 & 83.3 & 0.0 & 0.0 & 0.0 & 100.0 & 16.7 & 66.7 & 13.3 \\
 & \multirow[c]{2}{*}{OpenVLA-OFT} & SR & 66.7 & 83.3 & 16.7 & 50.0 & 50.0 & 100.0 & 0.0 & 100.0 & 0.0 & 33.3 & 26.7 & 73.3 \\
 & & F2 & 0.0 & 0.0 & 83.3 & 16.7 & 16.7 & 0.0 & 100.0 & 0.0 & 100.0 & 50.0 & 60.0 & 13.3 \\
 & \multirow[c]{2}{*}{GR00T} & SR & 50.0 & 50.0 & 0.0 & 100.0 & 0.0 & 100.0 & 0.0 & 66.7 & 0.0 & 100.0 & 10.0 & 83.3 \\
 & & F2 & 16.7 & 16.7 & 100.0 & 0.0 & 100.0 & 0.0 & 100.0 & 33.3 & 83.3 & 0.0 & 80.0 & 10.0 \\
\midrule
\multirow[c]{6}{*}{1.6} & \multirow[c]{2}{*}{$\pi_{0.5}$} & SR & 0.0 & 50.0 & 33.3 & 100.0 & 0.0 & 83.3 & 16.7 & 100.0 & 16.7 & 83.3 & 13.3 & 83.3 \\
 & & F2 & 100.0 & 50.0 & 66.7 & 0.0 & 83.3 & 0.0 & 50.0 & 0.0 & 83.3 & 16.7 & 76.7 & 13.3 \\
 & \multirow[c]{2}{*}{OpenVLA-OFT} & SR & 33.3 & 66.7 & 0.0 & 50.0 & 0.0 & 66.7 & 0.0 & 66.7 & 0.0 & 50.0 & 6.7 & 60.0 \\
 & & F2 & 0.0 & 0.0 & 100.0 & 50.0 & 100.0 & 16.7 & 83.3 & 0.0 & 100.0 & 50.0 & 76.7 & 23.3 \\
 & \multirow[c]{2}{*}{GR00T} & SR & 33.3 & 50.0 & 0.0 & 83.3 & 0.0 & 100.0 & 0.0 & 66.7 & 0.0 & 100.0 & 6.7 & 80.0 \\
 & & F2 & 0.0 & 16.7 & 100.0 & 16.7 & 100.0 & 0.0 & 100.0 & 33.3 & 100.0 & 0.0 & 80.0 & 13.3 \\
\bottomrule
\end{tabular*}
\end{table}

\begin{figure}[t]
\vspace{-10pt}
\centering
\begin{minipage}[b]{0.25\linewidth}
    \centering
    \includegraphics[width=\linewidth]{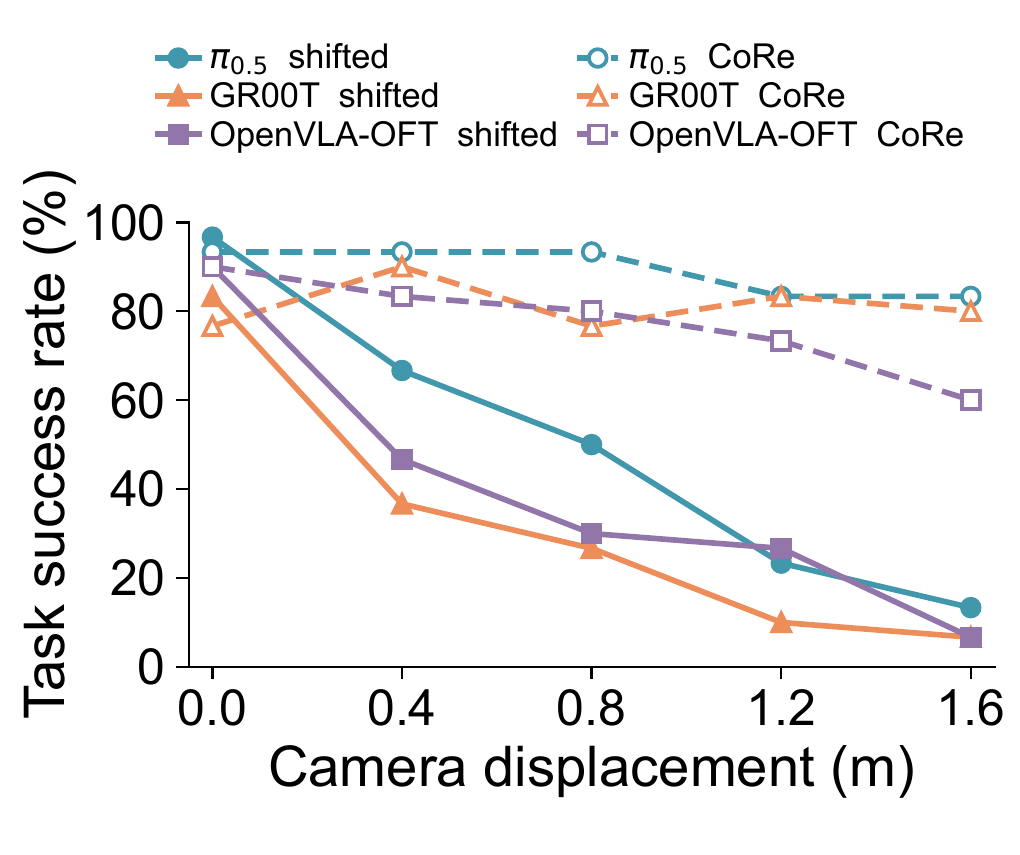}
    \par\vspace{2pt}
    {\small (a) Success rate}
\end{minipage}%
\begin{minipage}[b]{0.25\linewidth}
    \centering
    \includegraphics[width=\linewidth]{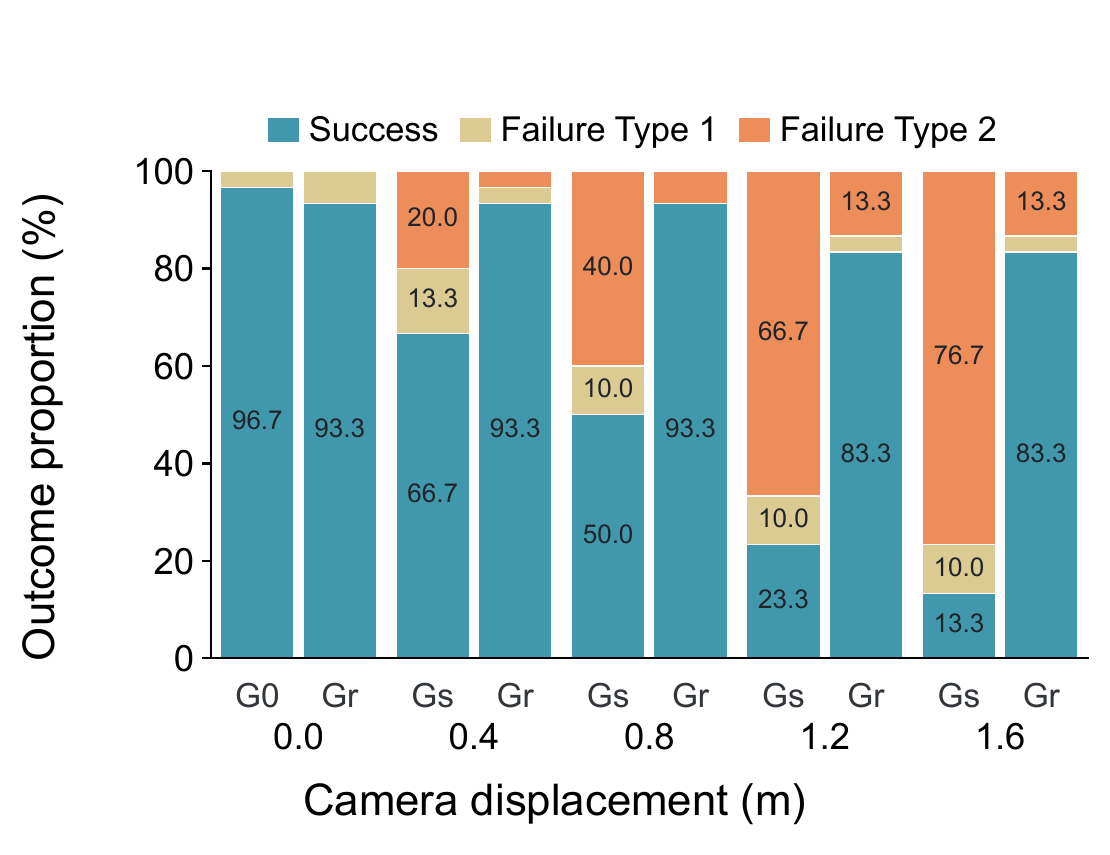}
    \par\vspace{2pt}
    {\small (b) $\pi_{0.5}$}
\end{minipage}%
\begin{minipage}[b]{0.25\linewidth}
    \centering
    \includegraphics[width=\linewidth]{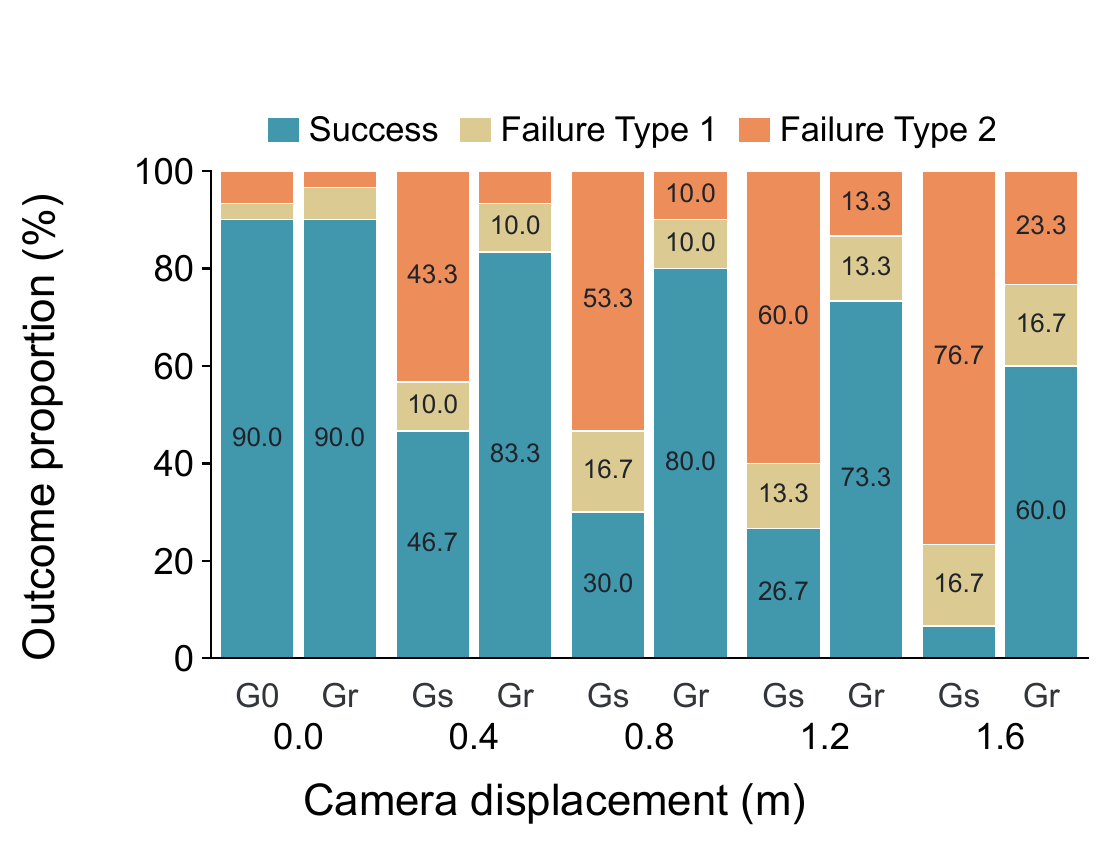}
    \par\vspace{2pt}
    {\small (c) OpenVLA-OFT}
\end{minipage}%
\begin{minipage}[b]{0.25\linewidth}
    \centering
    \includegraphics[width=\linewidth]{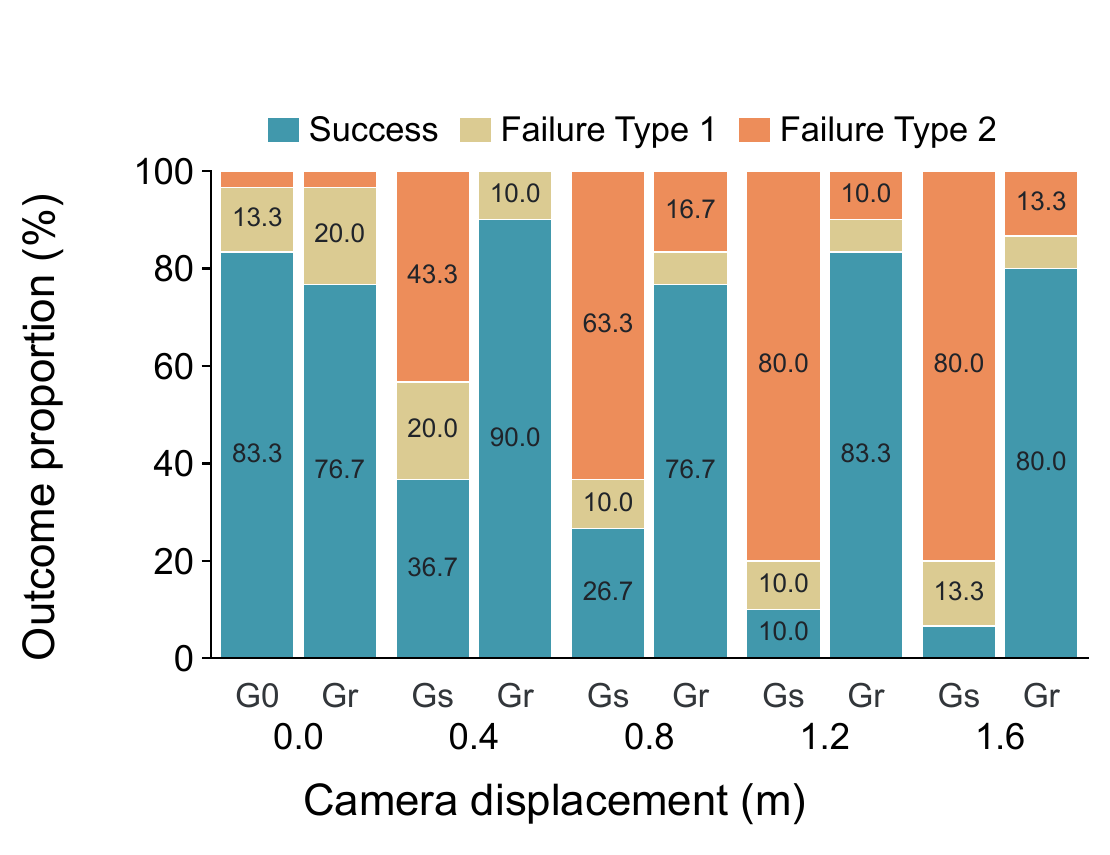}
    \par\vspace{2pt}
    {\small (d) GR00T}
\end{minipage}
\vspace{-20pt}
\caption{
Real-robot performance across 5 manipulation tasks under
external-camera shifts.
(a) Task success rates. 
(b--d) Outcome proportions for $\pi_{0.5}$, OpenVLA-OFT, and GR00T,
respectively.
Failure Types 1 and 2 denote local execution failure and
task-state inconsistency, respectively.
$G_0$ and $G_s$ denote raw external-camera
observations captured at the nominal and shifted positions,
respectively, while $G_r$ denotes observations transformed by CoRe-VLA.
}
\label{fig:main_results}
\end{figure}

\subsection{Evaluation in Real-robot Environment}
This experiment tests whether CoRe-VLA improves task success under external-camera shifts and reduces task-state inconsistency in real world environment.
We evaluate $\pi_{0.5}$, OpenVLA-OFT, and GR00T on the 5 real-robot tasks in Table~\ref{tab:real_robot_tasks}, with external-camera displacements of $0.0$ to $1.6$~m.
At each nonzero displacement, three camera positions are randomly selected, and each task is tested under two initial object layouts.
We compare raw external-camera observations with observations restored by CoRe-VLA.
The comparisons use identical camera positions, object layouts, and control settings, while wrist images are used without modification.
Success and the two failure categories defined in Sec.~\ref{sec:3} are recorded for each trial.
Table~\ref{tab:main_exp_detailed} reports per-task and overall success rates and Type 2 failure proportions under both observation conditions.
Figure~\ref{fig:main_results} further visualizes these results with success-rate curves and outcome-proportion plots, comparing raw observations with CoRe-VLA across camera displacements.
As shown in Fig.~\ref{fig:main_results}(a), success rates with raw observations decline sharply as displacement increases, whereas CoRe-VLA maintains substantially higher performance across the evaluated displacement levels.
At 1.6~m, CoRe-VLA increases the success rate of $\pi_{0.5}$ from 13.3\% to 83.3\%, and that of GR00T from 6.7\% to 80.0\%.
These results demonstrate that observation restoration can recover substantial task performance without adapting the VLA itself.
The outcome proportions in Fig.~\ref{fig:main_results}(b-d) further show that these improvements are accompanied by a marked reduction in task-state inconsistency (Type 2).
At 1.6~m, the proportion of all trials ending in Type 2 failures decreases from 76.7\% to 13.3\% for $\pi_{0.5}$ and from 80.0\% to 13.3\% for GR00T.
These reductions address the dominant failure pattern identified in Sec.~\ref{sec:3} and support the intended effect of CoRe-VLA: restoring the nominal external view for the VLA to prevent cross-view coordination breakdown. 
Qualitative examples and the corresponding analysis are shown in Fig.~\ref{fig:app_execution_examples} in Appendix.

\subsection{Evaluation on LIBERO}
\begin{wrapfigure}{r}{0.45\textwidth}
    \vspace{-40pt}
    \centering
    \includegraphics[width=\linewidth,trim=0 8 0 28,clip]{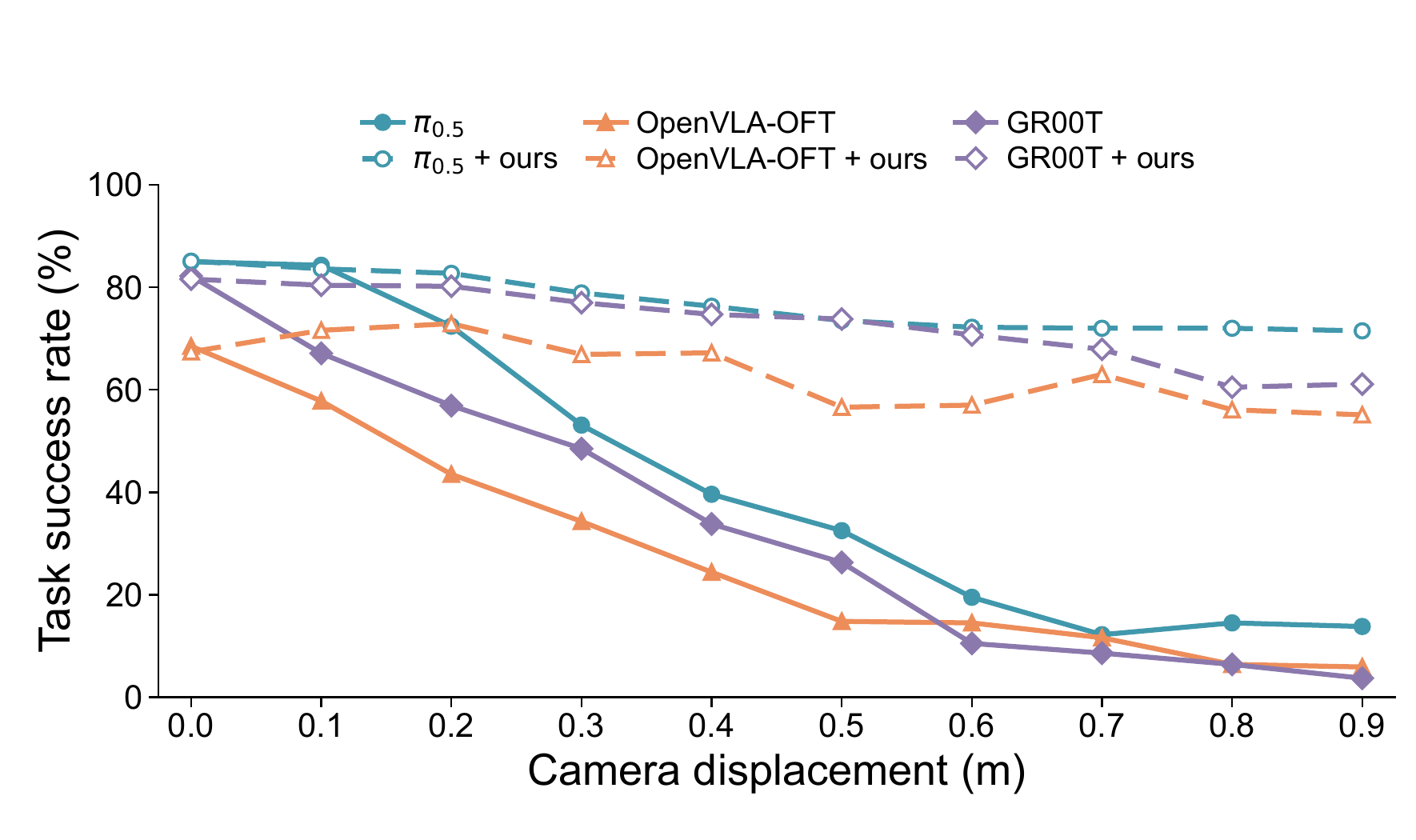}
    \caption{Task success rates on LIBERO-100 under external-camera shifts, without CoRe-VLA (solid) and with CoRe-VLA (dashed).}
    \label{fig:libero_results}
    \vspace{20pt}
\end{wrapfigure}

To evaluate whether the benefits observed in real-robot experiments extend to a broader task set, we conduct experiments on all 100 tasks in LIBERO~\citep{liu2023libero}.
External-camera displacement ranges from $0.0$ to $0.9$~m in increments of $0.1$~m, providing a finer-grained evaluation of viewpoint robustness.
Note that camera displacements of 1 m or greater are not evaluated due to the limited size of LIBERO scenes.
At each nonzero displacement, ten camera positions are randomly sampled.
We compare raw external observations with CoRe-VLA-restored observations, using identical camera poses, initial states, and random seeds.
The wrist observation remains unmodified, and the VLA parameters are kept frozen.

\begin{wraptable}{r}{0.45\textwidth}
    \vspace{-45pt}
    \vspace{-\intextsep}
    \centering
    \setlength{\abovecaptionskip}{2pt}
    \caption{Success rates (\%) on LIBERO-Plus Camera.}
    \label{tab:libero_plus_camera}
    \tiny
    \setlength{\tabcolsep}{2pt}
    \renewcommand{\arraystretch}{1.08}
    \begin{tabularx}{\linewidth}{@{}>{\raggedright\arraybackslash}X>{\raggedright\arraybackslash}p{0.30\linewidth}r@{}}
        \toprule
        Method & Backbone & SR (\%) $\uparrow$ \\
        \midrule
        \multicolumn{3}{@{}l}{VLA based methods} \\
        \midrule
        Cross-View AC~\citep{huang2026cross} & $\pi_{0.5}$ & 87.2 \\
        GAM~\citep{han2026geometric} & DA3-Giant & 83.1 \\
        Anchor-Align~\citep{dalal2026generalizable} & Prismatic-\newline Qwen2.5-0.5B & \textbf{96.3} \\
        AVA-VLA~\citep{xiao2026ava} & OpenVLA-OFT & 69.4 \\
        SRPO~\citep{fei2026srpo} & OpenVLA & 83.4 \\
        \addlinespace[2pt]
        $\pi_{0.5}$~\citep{intelligence2025pi} & $\pi_{0.5}$ & 72.0 \\
        $\pi_{0.5}$ + CoRe-VLA (Ours) & $\pi_{0.5}$ & \underline{89.1} \\
        \midrule
        \multicolumn{3}{@{}l}{WAM based methods} \\
        \midrule
        SCVC~\citep{huang2026selective} & Cosmos Policy & 76.0 \\
        Robust-WAM~\citep{yan2026robust} & GE-Act & 54.8 \\
        Robust-WAM~\citep{yan2026robust} & FastWAM & 28.7 \\
        \bottomrule
    \end{tabularx}
    \par\vspace{2pt}
    % \raggedright
    % $^{\dagger}$Held-out orbital extrapolation only, not the full Camera subset.
    % Cross-View AC masks wrist inputs; AVA-VLA uses per-suite policies; SRPO uses augmented data.
    \vspace{-20pt}
\end{wraptable}

As shown in Fig.~\ref{fig:libero_results}, CoRe-VLA substantially mitigates the performance degradation caused by increasing camera displacement for both $\pi_{0.5}$ and OpenVLA-OFT.
At $0.9$~m, it improves the success rate of $\pi_{0.5}$ from 13.8\% to 71.5\%, and that of OpenVLA-OFT from 5.9\% to 55.1\%.
Although performance still varies with displacement, the restored observations yield a much smaller overall decline than raw observations.
Together with the real-robot results, these findings support the effectiveness of CoRe-VLA across a broader collection of manipulation tasks and different VLA backbones, without additional multi-view data collection or VLA retraining.

\subsection{Evaluation on LIBERO-Plus}
\label{sec:libero_plus}
To test robustness under standardized camera shifts and compare CoRe-VLA with existing methods, we further evaluate CoRe-VLA on the camera subset of LIBERO-Plus~\citep{fei2026libero}.
In the expriment, The officially released $\pi_{0.5}$ weights are used and remain frozen throughout evaluation.
The external view is restored using simulator-provided depth and camera parameters, while the wrist observations remain unchanged.

Table~\ref{tab:libero_plus_camera} shows that CoRe-VLA improves the success rate of $\pi_{0.5}$ from 72.0\% to 89.1\%, a gain of 17.1 percentage points without additional multi-view data collection or VLA fine-tuning.
Although Anchor-Align reports a higher score, it adapts the VLA, whereas CoRe-VLA leaves the policy parameters unchanged.
The reported WAM results also indicate that world modeling alone does not guarantee robust execution under camera shifts.

\subsection{Ablation Study}

\begin{wrapfigure}{r}{0.35\linewidth}
    \vspace{-40pt}
    \centering
    \includegraphics[width=\linewidth]{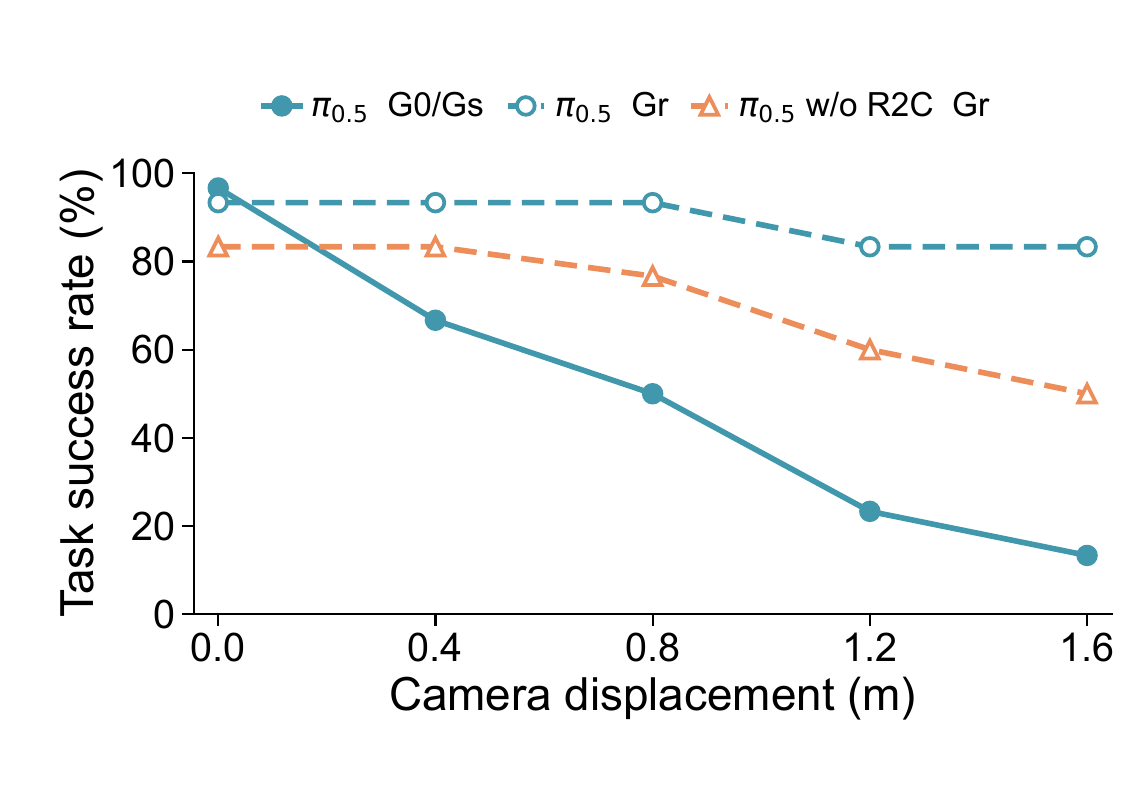}
    \vspace{-20pt}
    \caption{Effect of R2C Restoration on $\pi_{0.5}$ task success under external-camera shifts.}
    \label{fig:ablation_r2c}
    \vspace{-15pt}
\end{wrapfigure}

\textbf{Effect of R2C Restoration on execution outcomes. }
To isolate the contribution of R2C, we remove it from CoRe-VLA while retaining point-cloud reconstruction, re-rendering, and the frozen $\pi_{0.5}$ backbone.
All other experimental settings are identical to those described in Sec.~.
Figure~\ref{fig:ablation_r2c} compares this variant with full CoRe-VLA and direct use of the raw camera observations on the 5 real-robot tasks.
Even at zero displacement, success falls from 96.7\% with the raw image to 83.3\% without R2C, showing that re-rendering itself introduces a harmful appearance mismatch.
R2C raises success at this setting to 93.3\%, recovering most of the lost performance.
As displacement increases, newly exposed but unobserved regions can make the rendered image less complete and less sharp.
At 1.6\,m, success is 50.0\% without R2C but 83.3\% with it, a gain of 33.3 percentage points.
This improvement is consistent with R2C reducing rendering-induced texture loss, color distortion, and blur, making the restored external view more useful alongside the unchanged wrist view.

\begin{wraptable}{r}{0.45\textwidth}
\vspace{-20pt}
\centering
\caption{Effect of ETCA on execution efficiency.}
\label{tab:etca_time}
\begingroup
\tiny
\setlength{\tabcolsep}{2pt}
\renewcommand{\arraystretch}{1.15}
\begin{tabular}{@{}lrrrr@{}}
\toprule
Method & \shortstack{Latency\\(ms)} & \shortstack{Success\\(\%)} & \shortstack{Completion\\(s)} & \shortstack{Idle\\(s)} \\
\midrule
$\pi_{0.5}$ & 244 & 96.7 & 47.2 & 22.5 \\
$\pi_{0.5}$ + ETCA & 244 & 96.7 & 34.4 & 9.0 \\
$\pi_{0.5}$ + CoRe-VLA w/o ETCA & 478 & 93.3 & 64.5 & 39.1 \\
$\pi_{0.5}$ + CoRe-VLA & 478 & 93.3 & 43.7 & 13.8 \\
\bottomrule
\end{tabular}
\endgroup
\end{wraptable}

\noindent\textbf{Effect of ETCA on execution time. }
To assess whether ETCA mitigates the execution-time overhead of viewpoint restoration, four configurations are compared at the nominal camera position across our real-robot benchmark, with two initial layouts and three repetitions per layout.
All inference runs on an NVIDIA RTX 5060 Ti (16 GB), with other settings held constant.
Configurations without ETCA use synchronous execution.
Timing statistics include only matched trials completed successfully by all four configurations.
As shown in Table~\ref{tab:etca_time}, ETCA reduces CoRe-VLA's mean completion time from 64.5 to 43.7~s (32.2\%) and idle time from 39.1 to 13.8~s (64.7\%), despite unchanged single-inference latency.
It also reduces completion time for standalone $\pi_{0.5}$ by 27.1\%, with no observed success-rate decrease in either comparison.
These results support ETCA's effectiveness in reducing execution time through asynchronous inference and execution rather than faster individual inferences.
We further analyze how the individual loss terms affect R2C Restoration quality in Appendix~\ref{app:r2c_visual}.

\section{Conclusion and Limitations}
This work identifies \emph{cross-view coordination breakdown} in VLAs under camera shifts, which is the major cause of task failure.
The proposed CoRe-VLA demonstrates that this coordination breakdown can be mitigated without collecting additional data or fine-tuning the VLA: using point-cloud reconstruction and re-rendering to restore the joint observations familiar to the VLA during training.
Although extensive experiments show that CoRe-VLA substaintially improves VLA viewpoint robustness and ETCA can reduce task completion time, there are several limitation in this work.
First, CoRe-VLA relies on reasonably accurate camera depth measurements.
Additionally, Large viewpoint changes or occlusions can reveal regions absent from the source observation, which neither reprojection nor R2C can reliably recover.
Developing viewpoint robust methods that are independent of camera hardware and reliable under large camera displacements remains an important direction for future work.

\subsection*{AI use statement}
In this work, we used generative AI tools to polish manuscript language, preliminarily screen relevant literature, and discuss the feasibility of the method design. 
We did not use generative AI to implement CoRe-VLA or deploy the experiments. 
We have reviewed all AI-assisted material, including every AI-polished sentence, and verified the existence of related work. 
We take responsibility for the final content of this work, including text, claims, and artifacts produced with the aid of generative AI.

\bibliography{iclr2027_conference}
\bibliographystyle{iclr2027_conference}

\newpage
\appendix
\section*{Appendix}
\label{app:overview}
This appendix provides the success criteria for the five real-robot tasks (Appendix~\ref{app:success_criteria}) and additional details on the LIBERO benchmarks (Appendix~\ref{app:benchmarks}).
Appendix~\ref{app:r2c_pairs} is reserved for the construction of R2C training pairs, while Appendix~\ref{app:execution_examples} is reserved for analysis of qualitative execution examples.
Appendix~\ref{app:r2c_visual} presents a preliminary visual ablation of R2C, and Appendix~\ref{app:async_comparison} discusses related asynchronous execution methods

\section{Real-Robot Task Success Criteria}
\label{app:success_criteria}
Table~\ref{tab:app_success_criteria} specifies the success criteria for the tasks defined in Table~\ref{tab:real_robot_tasks}.
For sequential tasks, all required steps must be completed in the instructed order.

\begin{table}[b]
\centering
\caption{Success criteria for the five real-robot tasks.}
\label{tab:app_success_criteria}
\footnotesize
\setlength{\tabcolsep}{4pt}
\renewcommand{\arraystretch}{1.12}
\begin{tabularx}{\linewidth}{@{}cY@{}}
\toprule
\textbf{Task} & \textbf{Success criterion} \\
\midrule
T1 & The robot grasps the green glue stick and releases it inside the pink cup. \\
T2 & The robot places and releases the yellow tape measure inside the basket, then places and releases the pink cup inside the basket. \\
T3 & The robot first opens the lid so that it no longer covers the basket, then places and releases the yellow tape measure inside the basket. \\
T4 & The robot grasps the green glue stick and releases it on the side of the pink cup farther from the basket, within a $\pm45^\circ$ sector about the basket-to-cup direction. \\
T5 & The robot grasps the object initially closest to the basket and releases it inside the basket. \\
\bottomrule
\end{tabularx}
\end{table}

\section{Introduction of Simulation Benchmarks}
\label{app:benchmarks}
LIBERO~\citep{liu2023libero} is a simulated benchmark for language-conditioned robot manipulation, providing task instructions, proprioceptive states, and images from workspace and wrist cameras.
It includes suites focused on spatial relationships, object identity, and task goals, while LIBERO-100 contains 100 tasks involving diverse object interactions and motor skills.
LIBERO-Plus~\citep{fei2026libero} extends this benchmark with controlled perturbations across seven dimensions to assess model robustness beyond standard task completion.
Its Camera subset contains 1,599 instances across the Spatial, Object, Goal, and Long suites, with perturbations to camera distance, position, and orientation.

\section{Construction of R2C Training Pairs}
\label{app:r2c_pairs}
Training pairs are constructed from RGB frames in existing 2D task demonstrations, without depth maps, camera calibration, or additional views.
For each original frame $G_0$, a severity-controlled projective warp creates an image with uncovered regions.
Within the valid image region, short gaps are sampled near object contours, followed by sparse clusters of missing pixels; all uncovered and missing pixels are set to black.
The warp parameters and missing-pixel masks vary across frames, producing degraded inputs $\tilde{G}_0$ paired with their original targets $G_0$.
This procedure approximates the visible gaps caused by point-cloud re-rendering, rather than synthesizing a geometrically accurate new view.

\section{Qualitative Execution Examples}
\label{app:execution_examples}
\begin{figure}[t]
\centering
\includegraphics[width=0.95\linewidth]{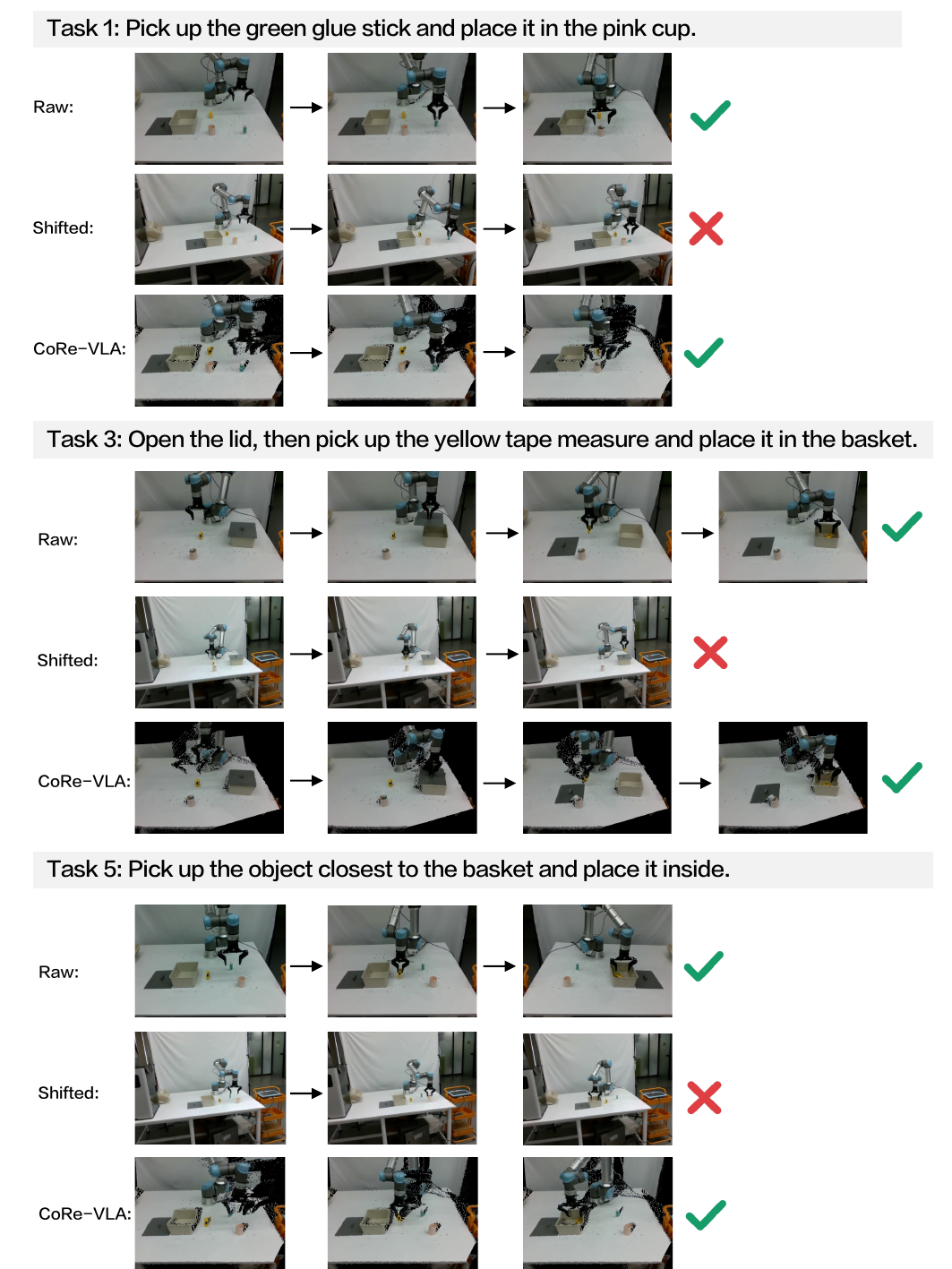}
\caption{Representative real-robot execution sequences under nominal (Raw), shifted, and CoRe-VLA-restored external views. Arrows indicate action order; check marks and crosses denote success and failure, respectively.}
\label{fig:app_execution_examples}
\end{figure}

Figure~\ref{fig:app_execution_examples} compares executions of Task 1, Task 3, and Task 5 under nominal, shifted, and CoRe-VLA-restored external views.
In Task 1, the shifted view causes a simple pick-and-place task to fail even though the relevant objects remain visible, while the nominal and restored-view executions succeed.
Task 3 and Task 5 more directly illustrate cross-view coordination breakdown.
In Task 3, the robot must open the lid before placing the tape measure in the basket, but the shifted-view execution acts on the locally visible tape measure without maintaining this order.
In Task 5, selecting the object closest to the basket requires the global spatial relationships provided by the external view.
Under the shifted view, the robot instead selects the cup visible in its wrist view.
These examples show that local cues can remain actionable while the global context needed for task-state interpretation and target selection is no longer used effectively.
By restoring an external view familiar to the VLA from training, CoRe-VLA helps align wrist-level cues with the intended task order and spatial constraints, despite residual artifacts in the restored images.

\section{Visual Ablation of R2C Restoration}
\label{app:r2c_visual}
\begin{figure}[t]
\centering
\includegraphics[width=\linewidth]{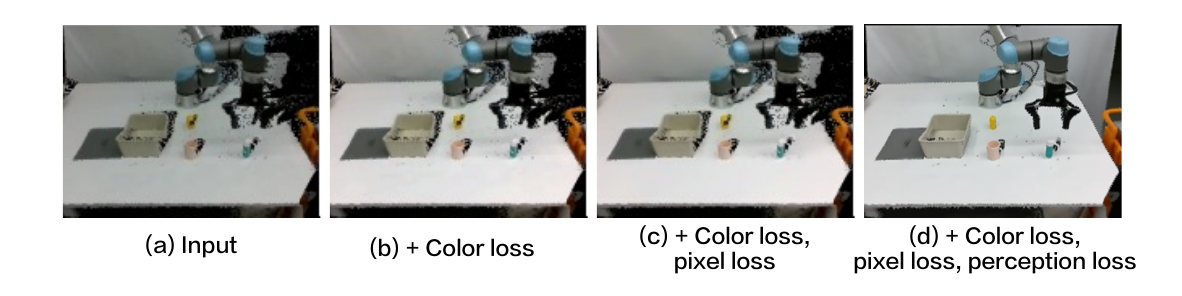}
\caption{Qualitative ablation of the R2C training losses. From left to right: original image, degraded input, and restorations using color loss, color and pixel losses, and the full objective including perceptual loss.}
\label{fig:r2c_loss_ablation}
\end{figure}

Figure~\ref{fig:r2c_loss_ablation} compares restorations from the same degraded input as the loss terms are added.
The input contains missing regions around the robot and basket, speckled artifacts, and unclear object boundaries.
Color loss alone, or combined with pixel loss, leaves much of this corruption visible.
Adding the perceptual loss produces the closest match to the original image in this example, with a more coherent robot appearance and clearer tabletop objects.
This visual comparison complements the execution-outcome ablation in Sec.~\ref{sec:r2c}.

\section{Comparison with Asynchronous Execution Methods}
\label{app:async_comparison}
Asynchronous execution allows the robot to act while the next action chunk is being inferred, but the incoming chunk must be reconciled with motion already performed.
Real-Time Chunking (RTC)~\citep{black2026real} addresses this for diffusion- and flow-based policies by freezing committed actions and inpainting the remaining chunk during sampling.
REMAC~\citep{wang2026real} learns corrective adjustments through masked action chunking to handle inconsistencies during asynchronous execution.
ETCA instead operates on an already predicted chunk: it matches candidate prefixes to the robot's measured joint trajectory and executes the corresponding suffix only when the match passes quality checks.
This makes ETCA particularly suited to CoRe-VLA's variable observation-processing latency without changing the frozen VLA or its action-generation procedure.

\end{document}